\documentclass[letterpaper, 10 pt, conference]{ieeeconf}  
\usepackage{color,soul}
\IEEEoverridecommandlockouts                              

\usepackage{amsmath} 
\usepackage{amssymb}  
\usepackage{cite}
\usepackage{graphicx, subcaption}
\usepackage{setspace}
\usepackage{mathrsfs}
\usepackage[ruled,vlined,titlenumbered]{algorithm2e}
\usepackage{balance}
\usepackage{multirow}
\usepackage{multicol}
\usepackage{makecell}
\usepackage{array, caption, tabularx,  ragged2e,  booktabs}
\usepackage{CJKutf8}
\PassOptionsToPackage{hyphens}{url}\usepackage{hyperref}  
\newcolumntype{P}[1]{>{\centering\arraybackslash}p{#1}}  
\newcolumntype{M}[1]{>{\centering\arraybackslash}m{#1}}  
\usepackage{fancyhdr}  

\newlength\mylen

\title{\LARGE \bf
Quick Learner Automated Vehicle Adapting its Roadmanship to Varying Traffic Cultures with Meta Reinforcement Learning}

\author{Songan Zhang$^{1}$, Lu Wen$^{1}$, Huei Peng$^{1}$, and H. Eric Tseng$^{2}$

\thanks{$^{1}$Songan Zhang, Lu Wen and Huei Peng are with the University of Michigan, Ann Arbor, MI 48109, USA. {\tt\footnotesize\{\small songanz, lulwen, hpeng\footnotesize\}\small@umich.edu}.}%
\thanks{$^{2}$H Eric Tseng is with Ford Motor Company.}%
}

\begin{document}

\maketitle
\thispagestyle{fancy}
\pagestyle{fancy}
\renewcommand{\footrulewidth}{0pt}
\renewcommand{\headrulewidth}{0pt}

\begin{abstract}
It is essential for an automated vehicle in the field to perform discretionary lane changes with appropriate roadmanship – driving safely and efficiently without annoying or endangering other road users – under a wide range of traffic cultures and driving conditions. While deep reinforcement learning methods have excelled in recent years and been applied to automated vehicle driving policy, there are concerns about their capability to quickly adapt to unseen traffic with new environment dynamics. We formulate this challenge as a multi-Markov Decision Processes (MDPs) adaptation problem and developed Meta Reinforcement Learning (MRL) driving policies to showcase their quick learning capability. Two types of distribution variation in  environments were designed and simulated to validate the fast adaptation capability of resulting MRL driving policies which significantly outperform a baseline RL.

\end{abstract}

\section{INTRODUCTION}
\label{INTRODUCTION}
In the last decade, Reinforcement Learning (RL) methods have excelled and been applied to many problems, including video games\cite{justesen2019deep}, robotic manipulations\cite{kober2013reinforcement}, natural language processing\cite{melis2017state}, business management\cite{silver2013concurrent}, healthcare\cite{miotto2018deep} and intelligent transportation systems\cite{bazzan2013introduction}. Amongst the recent work in the field of RL, there are two inspiring success stories. The first one is a system that plays video games at a superhuman level\cite{mnih2015human}. The second well-known success story is AlphaGo, which combines supervised learning and reinforcement learning\cite{silver2016mastering} and defeated one of the top rank human Go masters. A wide variety of RL applications can be found in the comprehensive survey by Li\cite{li2017deep}.

However, RL in real world applications have not been as successful due to several factors. 
1) Real-world applications usually do not have a unique and clearly-defined reward function. Unlike playing Atari video games or Go/Chess, most real-world cases neither have a single goal, nor have a unique reward function. Instead, designing the reward function to balance multiple sub-goals, or discovering an underlying reward function from demonstrations\cite{ng2000algorithms}, or learning without a reward function\cite{eysenbach2018diversity} all are possible first task to be solved.
2) For real-world safety critical problems such as automated driving systems and robot manipulators, learning from mistakes is not practical due to the high penalty (e.g. crashes) that comes with real-life blunders. Recently, several work formulated RL safety problems to train policies with guard rails including \cite{wen2020safe} that utilized Constrained MDPs (CMDPs) and \cite{nageshrao2019autonomous} that augmented Q learning with a short-horizon control embedding safety rules or safety agents \cite{nageshrao2020vehicle}. Another approach uses Lyapunov functions\cite{chow2018lyapunov}, following safe exploration strategies\cite{wachi2018safe}. 
3) Video games, board games, and other closed world environments have fixed environment dynamics in how they transition, albeit highly complex. Most real-world challenges, on the other hand, are ever evolving and will inevitably contain environments with distribution not present in the training data. Thus an RL policy trained in a single simulation environment may find it difficult to generalize towards the  real-word distribution. There are more challenges need to be addressed to implement RL to real-world problems, and we refer readers to the work by Gabriel\cite{dulac2019challenges}. 

In this paper, we will address the third gap in AV applications. Many challenges remain in capturing all distribution of the real world environment transition function into automated driving simulations. In addition to sensors and actuators uncertainties, environmental agents such as other road users are especially difficult to characterize, and all of them can drift overtime. In this paper, we will focus on the changing distribution of traffic environment, since it can be considered as the most challenging issue with human behaviors and driving styles being diverse and known to change overtime \cite{jaafra2019robust}.

Amongst all decision-making problems in autonomous driving, the lane change is an important feature for automated vehicles to maintain mobility. A lane change that is not urgent but desirable (e.g. overtaking a slower vehicle) is considered a Discretionary Lane Change (DLC), as opposed to Mandatory Lane Change (e.g. required upon lane closing). In this work, we formulate the discretionary lane change problem as a quasi-static MDP, and the automated vehicle is supposed to learn to adapt to a range of different MDPs.

Meta-RL-based policies for discretionary lane change applications were developed in this paper. To validate our method, we experiment with two types of varying environment distributions. The first considers varying distribution/number of adversarial vehicles, and the second contains varying distribution of surrounding driving behaviors - ranging from egoistic to altruistic by assigning different parameters in the driver model. We want to validate that the trained policy can handle both hostile and normal (albeit varying) surrounding vehicles. The rest of this paper is organized as follows. Problem definition and literature are shown in Section II. The background knowledge is described in Section III. In Section IV the simulation environments are interpreted, followed by the training and testing setup in Section V. Simulations results are shown and analyzed in Section VI. Finally the paper is concluded with Section VII.

\section{RELATED WORKS}
\label{RELATED WORKS}
Different methods were used to design fast learning agents to adapt to unseen environments.  Markov Decision Process (MDP) is a popular approach for such design problems.  Approaches using multi-MDPs adopt different optimization and adaptation methods. Frequently, the agent trains an identifier using supervised learning. Subsequently, for each identified model, Model Predictive Control (MPC) or Dynamic Programming (DP) methods can be used to learn the policy. During adaptation, we can witch to the corresponding controller based on the identified model. For instance, in \cite{nagabandi2018learning}, Nagabandi et al. use meta learning to train a dynamics model prior. And this prior can rapidly adapt to the local context when combined with recent data. The controller is extracted using model predictive path integral control. However, the models need to be enumerated with their structure, limiting the agent’s generalization ability.

In other studies, researchers use behavior cloning for the adaptation step. For example, in \cite{finn2017one, yu2018one}, the authors present Domain-Adaptive Meta-Learning, a system that allows robots to learn from a single video of a human via prior meta-training data collected from related tasks. During training, the agent is provided with demonstration data. The agent is taught how to infer a policy from just one demonstration. During testing, only one expert demonstration is provided, and the agent runs behavior cloning. The performance after the adaptation can be outstanding, however, it requires expert demonstration and in our work, we don’t assume to have it. 

Model-free Meta Reinforcement Learning (MRL) can also solve adaptation problems. In \cite{duan2016rl, wang2016learning, heess2015memory} the authors use a Recurrent Neural Network (RNN) and encode the MDP’s information as the hidden memory of the RNN. And the policy contains the information in its weights to adapt to different environments. However, there is no mathematical convergence proof and we cannot guarantee that the RNN-based MRL methods adapt well or converge at all. Therefore, a more consistent MRL method is needed. 

Another class of MRL methods uses the policy gradient approach for both the meta training and the adaptation steps \cite{finn2017model, rothfuss2018promp, gupta2018meta}. In \cite{finn2017model}, Finn et al. developed the Model Agnostic Meta-Learning (MAML) method. The idea is that the agent is trying to find the parameter $\theta$, such that when the agent takes a few gradient steps on that $\theta$, it will get to a $\theta_i^*$ which is optimal for a given MDP $\mathcal{M}_i$. However, policy gradient methods suffer in sparse reward environments. The agent cannot update its policy using trajectories with no reward. Also, if the reward functions are the same for different environments, the MAML agent may not capture the environments' features. 

Both the RNN-based and gradient-based approaches use on-policy RL methods for both the meta training and the adaptation steps and thus are data inefficient. The adaptation step is inherently on-policy learning since, given a new environment, the agent needs to collect new data using the current policy. On the other hand, the meta training step does not have to be on-policy. Leveraging a stochastic encoder to capture the context of adaptation data, Rakelly et al. \cite{rakelly2019efficient} developed an off-policy MRL with its meta training step based on the Soft Actor-Critic (SAC) \cite{haarnoja2018soft}. The developed method is called the Probabilistic Embedding for Actor-critic RL (PEARL). The PEARL MRL method is consistent, data-efficient, and has an advanced exploration strategy. 

In this work, we will implement the gradient based MAML method and the stochastic encoder based method PEARL in our DLC application and compare their performance.  

\section{META REINFORCEMENT LEARNING BASICS}
\label{PRELIMINARIES}
In this section, we will introduce the Meta Reinforcement Learning (MRL) basics and its notations. Unlike traditional Reinforcement Learning (RL) methods, in which a single Markov decision process (MDP) problem ($\mathcal{M}$) is solved, MRL tries to solve the multi-MDPs adaptation problem. The training and testing tasks of MRL are different but drawn from the same task distribution $p(\mathcal{M})$, where each task is a MDP, consisting of a set of states, actions, a transition function, and a bounded reward function. Each task $\mathcal{M}_i$ consists of the state space, the action space, the transition probability, the reward function and the discounted faction, i.e., $\mathcal{M}_i=(S_i, A_i, P_i, r_i, \gamma_i)$.

In the traditional RL, we solve:

\begin{equation}
    \alpha = \arg\max_{\alpha}\mathbf{E}_{\pi_{\alpha},\mathcal{M}}\left[\sum_{t=0}^{\infty}\gamma^tr(t)\right]
\end{equation}
where $\alpha$ is the parameter vector of policy $\pi$, $r$ is the reward function, and $\gamma$ is the discount factor. While in MRL, we solve:

\begin{equation}
\label{eq:meta RL}
    \theta = \arg\max_{\theta}\sum_{i=1}^{N}\mathbf{E}_{\pi_{\phi_i},\mathcal{M}_i}\left[\sum_{t=0}^{\infty}\gamma_i^t r_i(t)\right]
\end{equation}

\begin{equation}
\label{eq:meta RL_adaptation}
    \phi_i= f_\theta(\mathcal{M}_i)
\end{equation}
where the $\theta$ is the parameter of the adaptation function $f_\theta(\mathcal{M}_i)$. $\mathcal{M}_i$ the $i^\text{th}$ MDP environment. The meta-learner Equation (\ref{eq:meta RL}) adapts its parameter based on the sum of expected return collected from performing each task with the adapted policy Equation (\ref{eq:meta RL_adaptation}). This meta learning step of Equation (\ref{eq:meta RL}) is understood as the outer loop, whereas the adaptation step Equation (\ref{eq:meta RL_adaptation}) is the inner loop that fine tunes its parameter based on trajectories collected with the initial policy associated with the meta learner parameter.

In this work, we will implement two state-of-the-art MRL methods which are the MAML \cite{finn2017model} and PEARL \cite{rakelly2019efficient} method. In the MAML approach, both the meta training and adaptation loops are performed using policy gradient, for which the updating equation of the meta training step, integrated with parameter update of inner loops, is: 
\begin{equation}
    \theta \leftarrow \theta + \alpha_1 \sum_i \bigtriangledown_\theta r_i[\theta + \alpha_2 \bigtriangledown_\theta r_i(\theta)]
    \label{Eq:MAML}
\end{equation}
where $\alpha_1$ and $\alpha_2$ are the learning rates, and $r_i$ is the reward function of the $i^\text{th}$ MDP. After the meta training step, the agent will learn a $\theta$ that is sensitive to all given reward functions. So that starting from that $\theta$, with only a few steps of adaptation, the agent can find a better policy.

\begin{figure}[ht]
\centering
	\includegraphics[width=0.48\textwidth]{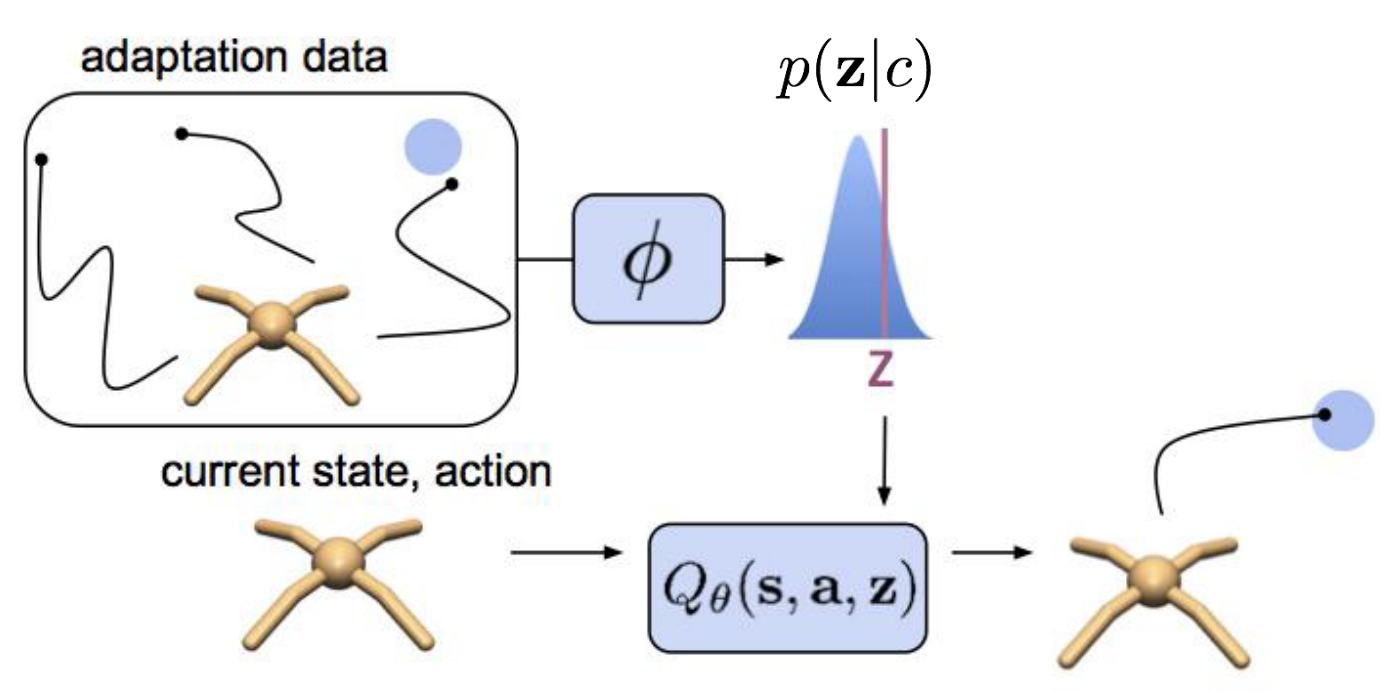}
	\caption{PEARL method illustration. Image is from \cite{rakelly2019efficient}.}
	\label{fig:PEARL}
\end{figure}

In the PEARL method \cite{rakelly2019efficient} shown in Figure \ref{fig:PEARL}, the agent consists of a stochastic encoder for adaptation and an off-policy RL SAC algorithm for meta learning.  During meta training, the encoder characterizes different environment with a latent variable $\mathbf{z}$ and form its belief of $p(\mathbf{z}|c)$, the $\mathbf{z}$ distribution given the context c, i.e. the batches of $(s, a, s', r)$ or adaptation data collected from the environment; while the SAC algorithm improves its policy given the belief (or specifically, given the latent variable $\mathbf{z}$ sampled from the belief of $p(\mathbf{z}|c)$) and feeds back its critic loss to coach the encoder. At inference, the agent collected recent data (of current environment) with initial policy to infer the belief, sampled from it, and feed it into the trained off-policy SAC.

In this work, we will implement both PEARL and MAML in our application. 



\section{VARYING TRAFFIC CULTURES}
\label{ENVIRONMENT DISTRIBUTIONS}
In order for an automated vehicle to minimize traveling time and avoid lanes with traffic shockwaves, performing discretionary lane change is necessary. For an automated vehicle, we further require it to drive with roadmanship, i.e., to make efficient lane changes without annoying or endangering other drivers, and to respond to crash threats safely – swiftly yet appropriately – without creating hazards for others. It is worth noting that while driving cautiously is usually safe, being overcautious is not acceptable \cite{AVstatistics}. 

Next section describes different traffic cultures that we will introduce on a simulated highway where we expect an MRL automated vehicle can quickly adapt to the culture and exhibit appropriate roadmanship accordingly. The different traffic cultures correspond to different MDP or statistical distribution of how the environment may transition.

\subsection{Varying adversarial vehicles (w.r.t. trained distribution)}
\label{Environments Attackers}
  
\begin{figure}[ht]
\centering
	\includegraphics[width=0.48\textwidth]{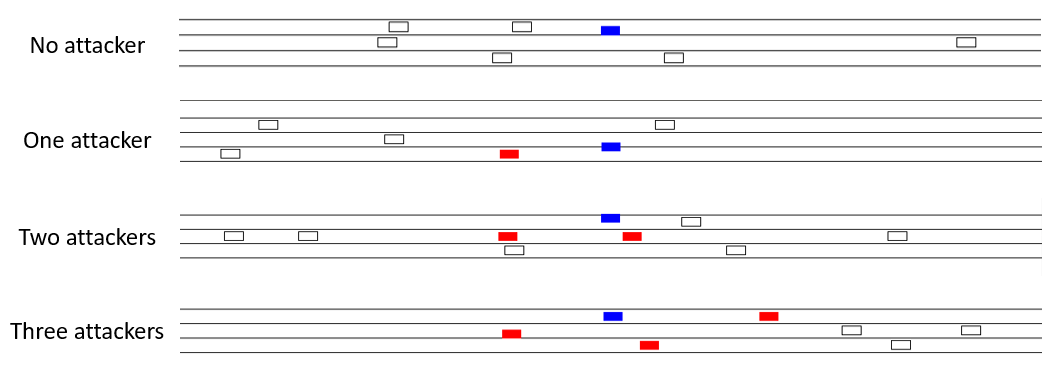}
	\caption{Examples of environments with different numbers of attackers. The blue car is the agent, the red cars are the attackers designed in \cite{Zhang_2020_CVPR_Workshops}, and the white cars are regular drivers designed in \cite{nageshrao2019autonomous}.}
	\label{fig:attacker_env}
\end{figure}

In \cite{Zhang_2020_CVPR_Workshops}, we generated "socially acceptable" attacks by training an adversarial attacker to explore the weakness of an AV with a fixed policy. The attacker attempted to invoke out of distribution traffic behaviors to confuse the AV. It showed that the trained attacker was capable of exploiting the fixed policy and induced collisions for which the AV is largely to blame. In this paper, we want to show that by implementing MRL, the trained AV agent can adapt to different environments and reduce crash rate. To demonstrate that, we design the distribution with three variables in Equation (\ref{equ:attacker_env}) to characterize the environments: 1) the traffic density variable $\alpha_{den}$, which is a scale of average distance between vehicles; 2) the number of total vehicles $n_{car}$, which can be sampled from 10 to 30; and 3) the number of attackers $n_{att}$, which is from 0 to 3. The attackers are randomly positioned around the AV.

\begin{equation}
    \begin{split}
        M_i = & \{\alpha_{den}, n_{car}, n_{att}\}, \alpha_{den}\sim U\left(0.5, 1.5\right), \\
        & n_{car}\sim U\{10, 30\}, n_{att}\sim U\{0,3\}
    \end{split}
    \label{equ:attacker_env}
\end{equation}
These variables determine the initial conditions of the simulation environment, as shown in Figure \ref{fig:attacker_env}. The reward functions for different environments are the same as in \cite{nageshrao2019autonomous}.

\subsection{Varying social behaviors from egoistic to altruistic}
\label{Environments IDM-Mobil}

In this experiment, we build the distribution of environments based on the highway-env \cite{highway-env} environment.

\begin{figure}[ht]
    \centering
    \includegraphics[width=0.48\textwidth]{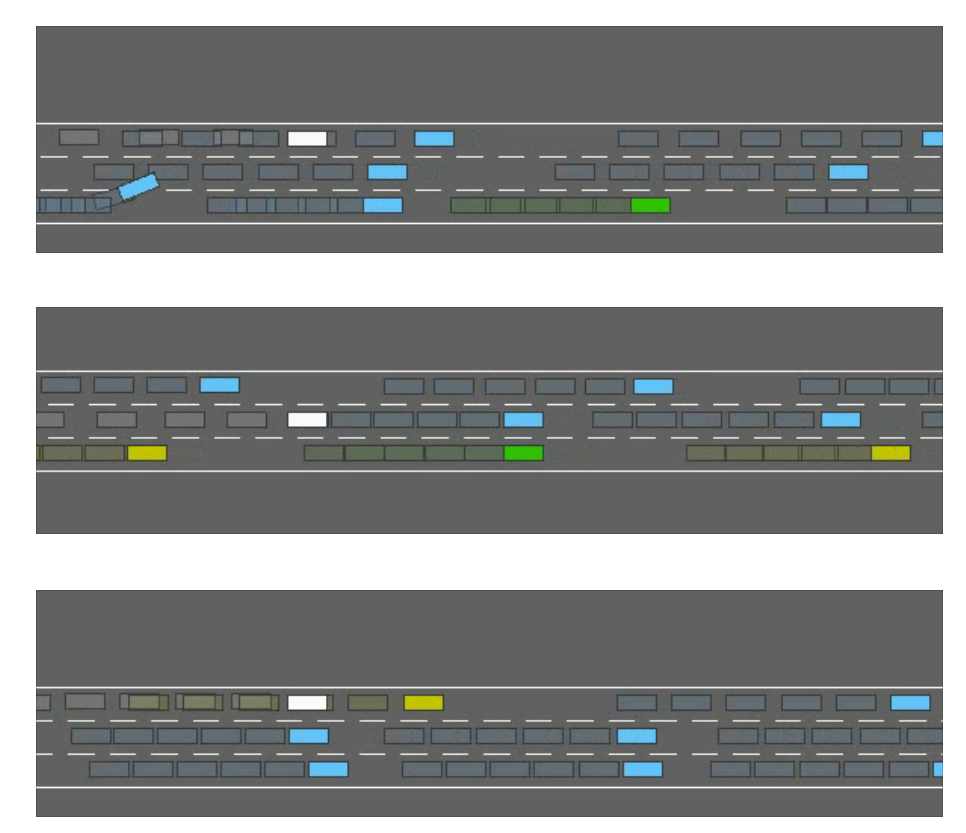}
    \caption{The highway-env environment \cite{highway-env}. The white box is the agent, the yellow boxes are aggressive drivers, the blue boxes are normal drivers and the green boxes are conservative drivers.}
    \label{fig:IDM_env}
\end{figure}

The state-space $S\subseteq R^n$ of the learning agent (the green box in Figure \ref{fig:IDM_env}) includes the host vehicle's lateral position $y$ and longitudinal velocity $v_x$, the relative longitudinal position of the $i^\text{th}$ surrounding vehicle $\Delta x^i$, the relative lateral position of the $i^\text{th}$ surrounding vehicle $\Delta y^i$ and its relative longitudinal velocity $\Delta v_x^i$. In total, we have a continuous state space of $2+3\times 6=20$ dimensions, i.e., $S\subseteq R^{20}$. The actions of the learning agent are the steering angle and acceleration, which are both continuous. The steering angle’s range is $[-\frac{\pi}{4}, \frac{\pi}{4}]$, and the acceleration’s range is $[-6\ m/s^2, 6\ m/s^2]$.

In the highway-env \cite{highway-env} environment, the surrounding vehicles are controlled by the IDM-Mobil model, and the vehicle will change lane when:

\begin{equation}
    \tilde{a}_c-a_c+p\left[(\tilde{a}_n-a_n)+(\tilde{a}_o-a_o)\right]>\Delta a_{th},
\end{equation}
and
\begin{equation}
    \tilde{a}_n>-b_{safe},
    \label{eq:safety_criterion}
\end{equation}
where $a_c$ is the ego vehicle’s acceleration in the current lane and $\tilde{a}_c$ is the potential ego vehicle’s acceleration if it changes lane. New and old successors are denoted as $n$ and $o$, the corresponding $a$ is the current acceleration and the $\tilde{a}$ is the potential if the ego vehicle changes lane. $p$ is the politeness factor and $\Delta a_{th}$ is the switching threshold. Therefore, the social behavior and aggressiveness of the surrounding vehicle can be represented by parameter $p$ and $\Delta a_{th}$. The Equation (\ref{eq:safety_criterion}) is the safety criterion that guarantees after the lane change, the deceleration of the successor in the target lane does not exceed a given safe limit $b_{safe}$. 

Since the politeness factor and the switching threshold are correlated for one kind of driver behavior, we do not sample them separately. Instead, we designed three different kinds of driver behavior: the aggressive driver, the normal driver, and the conservative driver \cite{schwarting2019social}. The corresponding parameters are listed in Table \ref{table:drivers}. From the table, we can see that the aggressive drivers will not consider other surrounding vehicles and may change lanes with a small acceleration advantage, while the conservative drivers do consider other surrounding vehicles and will change lanes only when there is a big acceleration advantage. The normal drivers are just in between.

\begin{table}[ht]
    \caption{Mobil Parameters for Different Driver Behaviors}
    \label{table:drivers}
	\centering
    \begin{tabular}{c|M{0.2\linewidth}|M{0.2\linewidth}|M{0.2\linewidth}}
        \hline
         Parameters & Aggressive Driver  & Normal Driver & Conservative Driver\\
         \hline
         \hline
         $p$ & $0$ & $0.3$ & $0.5$\\
         \hline
         $\Delta a_{th}$ & $0.8\ m/s^2$ & $1\ m/s^2$ & $1.2\ m/s^2$\\
         \hline
         $b_{safe}$ & $2\ m/s^2$ & $1\ m/s^2$ & $0.5\ m/s^2$\\
         \hline
    \end{tabular}
\end{table}

Each environment is decided by the following variables: the traffic density variable $\alpha_{den}$, which is a scale of the average distance between vehicles; the total number of vehicles $n$ which is the sum of the number of aggressive drivers $n_{agg}$, the number of normal drivers $n_{nor}$ and the number of conservative drivers $n_{con}$. To sample an environment, we first uniformly sample the traffic density variable $\alpha_{den}$ from 0.5 to 1.5 and the total number of vehicles $n$ from 10 to 30. Then the numbers of different driver behaviors (i.e. $n_{agg}$, $n_{nor}$ and $n_{con}$) are sampled from the multinomial distribution $M(n, k)$, where $n$ is the total number of vehicle and $k=\frac{1}{3}$. By sampling from $M(n, k)$, we will have $n_{agg}+n_{nor}+n_{con}=n$ and the probability of sampling from each category is the same.

The reward functions \cite{highway-env} for different environments are the same and is composed of a velocity term and collision term:
\begin{equation}
    R(s,a)=\alpha\left(\frac{v-v_{min}}{v_{max}-v_{min}}\right)-\beta r_{collision},
\end{equation}
where $v$, $v_{min}$, and $v_{max}$ are the current, minimum, and maximum speed of the agent, respectively, and $\alpha$, $\beta$ are weighting coefficients. For details, please refer to \cite{highway-env}.


\section{Training and Testing Setup}
\label{Baselines}
We have implemented the PEARL and MAML MRL methods in both environments with varying distributions and compared their meta training process in Figure \ref{fig:attacker_env_r} and Figure \ref{fig:IDM_env_r}. Both algorithms are trained in 8 different tasks sampled from each environments distribution. Then at the end of each iteration, the meta agent is tested in 4 unseen tasks sampled from each environments distribution to show the adaptation method. The average returns of 4 unseen tasks during meta training are compared with the x-axis being the total environment transition steps which represent the amount of data used to train the meta learner in Figure \ref{fig:attacker_env_r} and Figure \ref{fig:IDM_env_r}. And the returns of each algorithm are averaged across five random runs. Both the hyper-parameters of the PEARL and MAML are tuned using optuna package \cite{akiba2019optuna}, which is an open-source hyperparameter optimization framework to automate hyper-parameter search.

For evaluating the performance of trained meta agents, we compare the PEARL trained agent and the MAML trained agent with a fine-tune method based on the Trust Region Policy Optimization (TRPO) \cite{schulman2015trust} method with safety check implemented from \cite{zhang2019discretionary}. The fine-tune method will just keep updating the initial policy in a new environment given collected data. The evaluation results will be compared with the x-axis being the data collected in the new environment. We will sample $10^4$ different environments and evaluate all three approaches. 

For the DLC application, the crash rate of the trained agent is another important metric. Therefore, we will also report the crash rate of the fine-tune agent, MAML agent, and the PEARL agent in Section \ref{RESULTS}. We will compare the crash rate with the benchmark policy trained in the original environment using the method designed in \cite{zhang2019discretionary}.

\section{RESULTS}
\label{RESULTS}
\subsection{Training Results}

\begin{figure*}[ht]
	\centering
	\begin{subfigure}[b]{0.32\textwidth}
		\includegraphics[width=\textwidth]{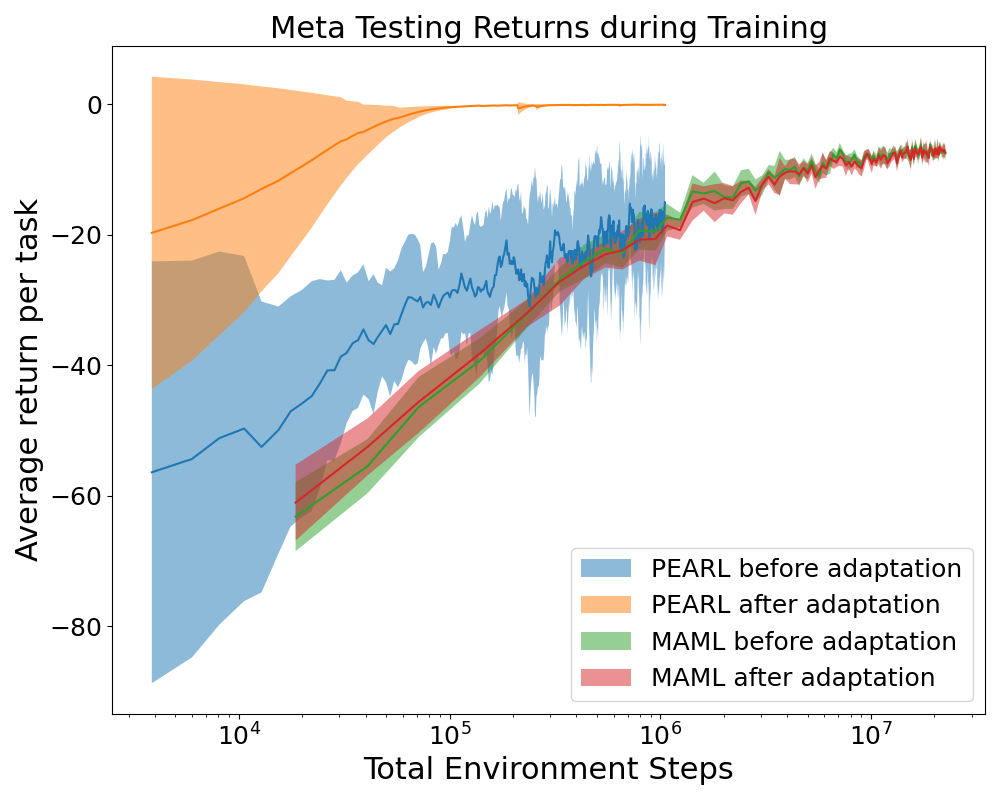}
		\caption{Meta Training}
		\label{fig:attacker_env_r}
	\end{subfigure}%
	\begin{subfigure}[b]{0.32\textwidth}
		\includegraphics[width=\textwidth]{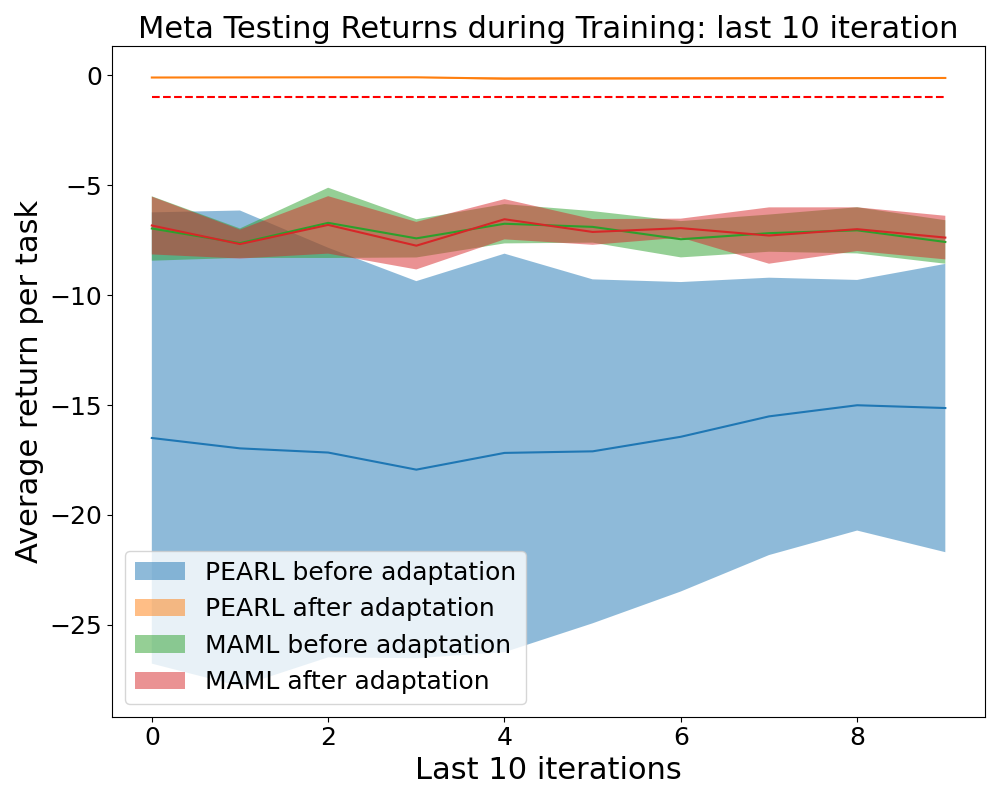}
		\caption{Meta Training: last 10 iterations}
		\label{fig:attacker_env_r_10}
	\end{subfigure}
	\begin{subfigure}[b]{0.32\textwidth}
		\includegraphics[width=\textwidth]{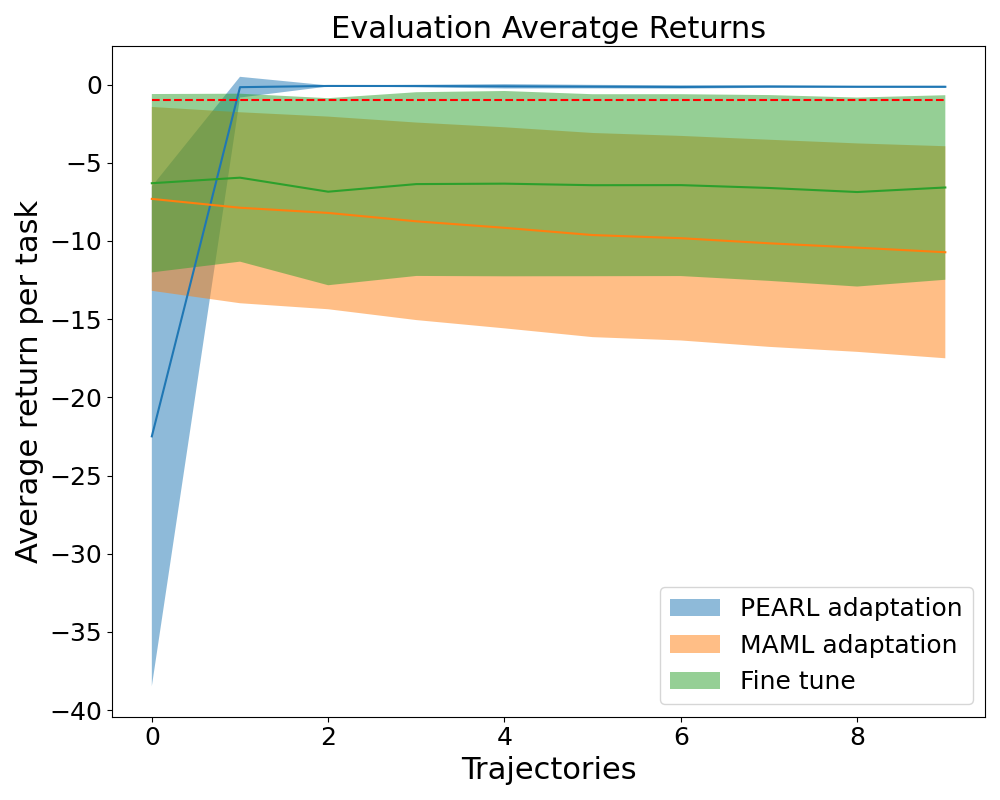}
		\caption{Evaluation Average Returns}
		\label{fig:attacker_env_r_eval}
	\end{subfigure}
	\caption{Meta Training and Evaluation Results in Attacker Environments}
\end{figure*}

\begin{figure*}[ht]
	\centering
	\begin{subfigure}[b]{0.32\textwidth}
		\includegraphics[width=\textwidth]{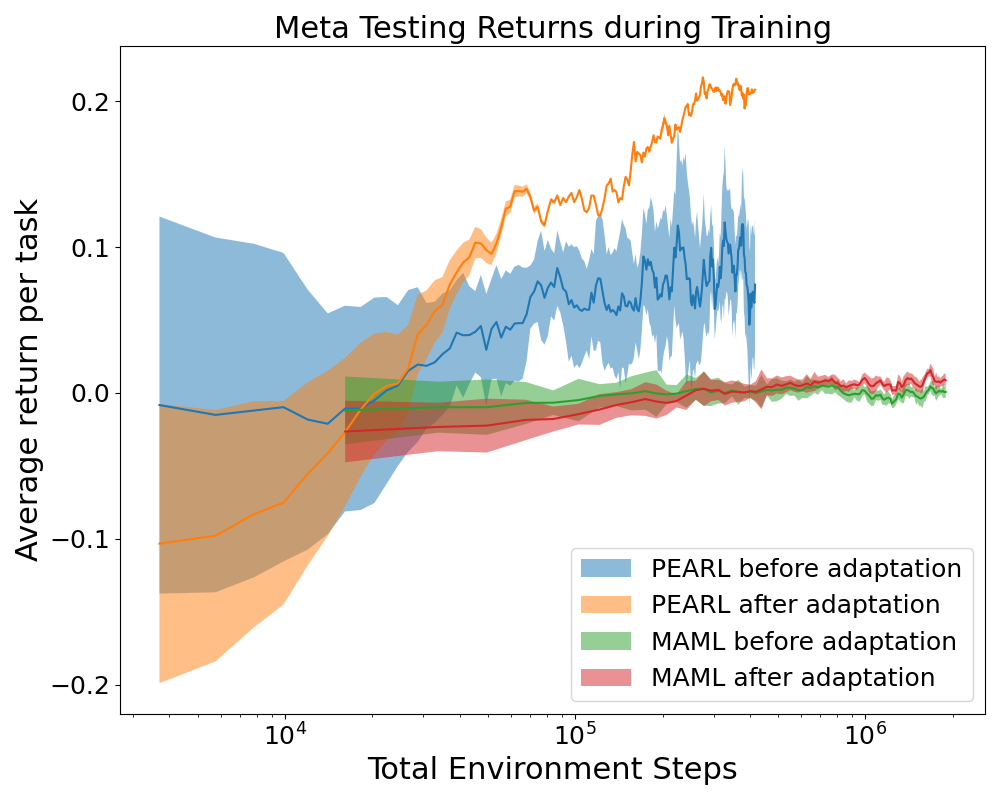}
		\caption{Meta Training}
		\label{fig:IDM_env_r}
	\end{subfigure}%
	\begin{subfigure}[b]{0.32\textwidth}
		\includegraphics[width=\textwidth]{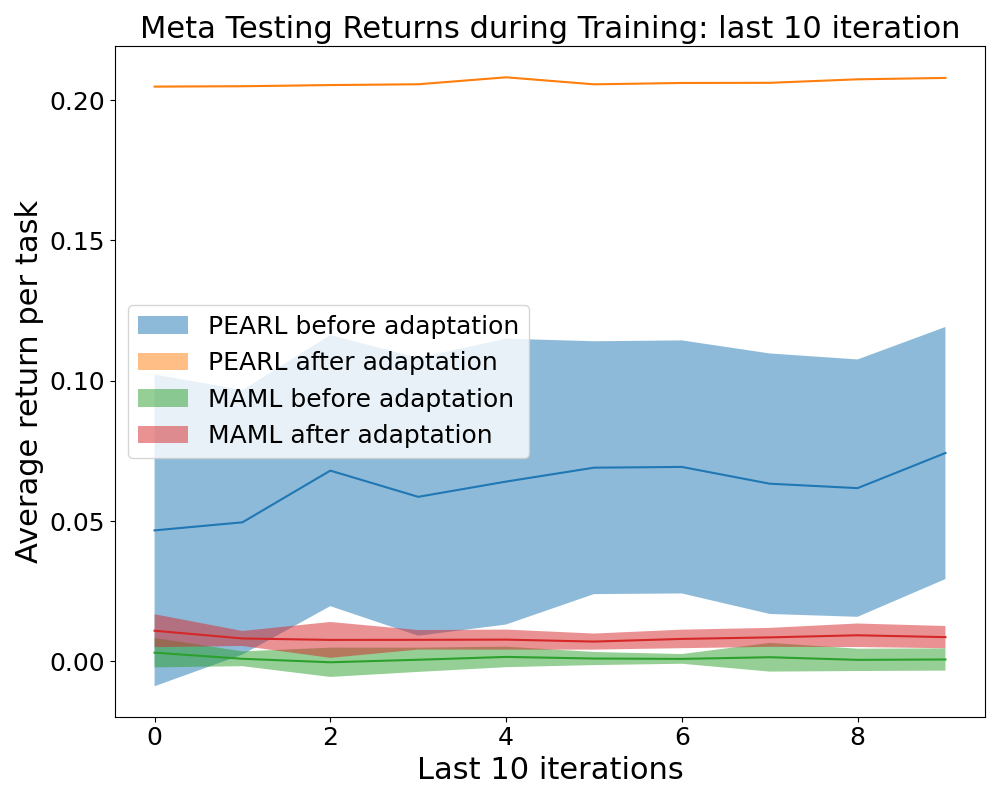}
		\caption{Meta Training: last 10 iterations}
		\label{fig:IDM_env_r_10}
	\end{subfigure}
	\begin{subfigure}[b]{0.32\textwidth}
		\includegraphics[width=\textwidth]{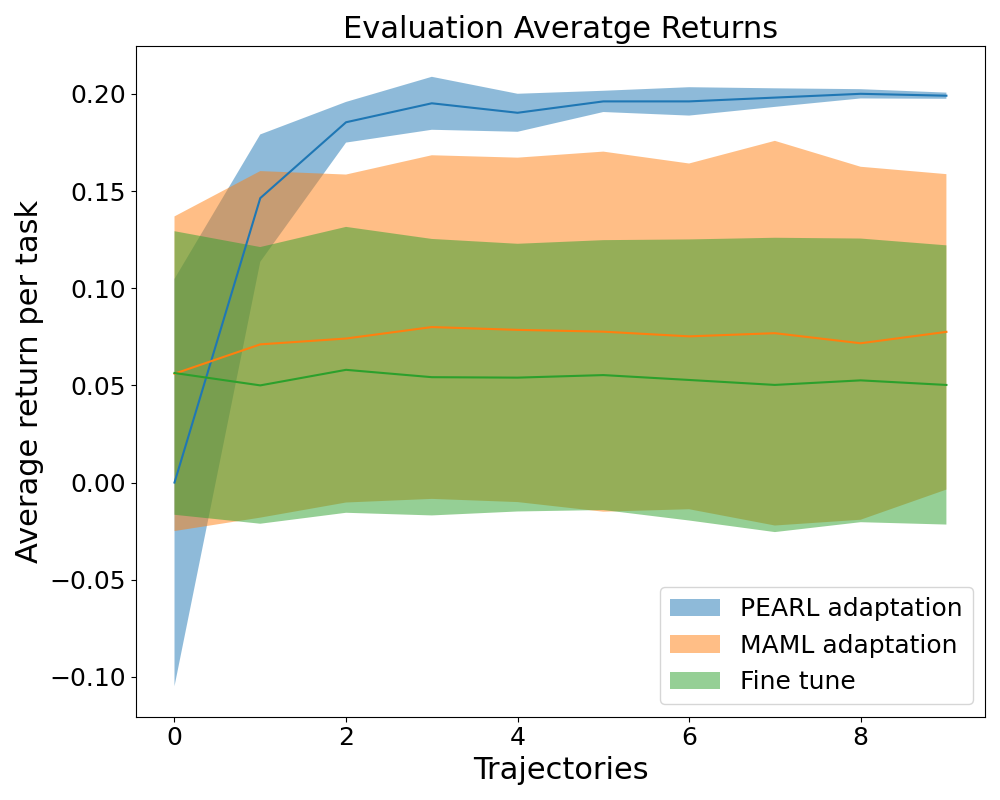}
		\caption{Evaluation Average Returns}
		\label{fig:IDM_env_r_eval}
	\end{subfigure}
	\caption{Meta Training and Evaluation Results in IDM-Mobil Environments}
\end{figure*}

This section shows the meta testing returns of the PEARL method and MAML method during the meta training. Results for the attacker environments distribution described in Section \ref{Environments Attackers} are shown in Figure \ref{fig:attacker_env_r} and Figure \ref{fig:attacker_env_r_10}. In Figure \ref{fig:attacker_env_r}, we show the before and after adaptation of PEARL and MAML in the logarithmic axis. The x-axis is the total environment steps $(s, a, s', r)$ representing how much data they use for training. As can be seen from the figure, the PEARL method converges after collecting $10^5$ data points, meanwhile the MAML converge after collecting $10^7$ data points. PEARL is one hundred times more data-efficient than MAML which is consistent with the conclusion in \cite{rakelly2019efficient}. Moreover, if we look at the before and after adaptation curve of each approach, we can see that the agent trained by the PEARL method shows good adaptation. While for the MAML method, there is almost no adaptation.

If we zoom in on the last ten iterations of MAML and PEARL and put them together, we can have this Figure \ref{fig:attacker_env_r_10}. The red dashed line is the crash line. The average reward below this line indicates there are crashes in that iteration. As you can see, the MAML not only shows no adaptation, but there are also still many crashes at the end of the training. While for PEARL, we can see that there is no crash after adaptation at the end of training. 

Results for the IDM-Mobil environment described in Section \ref{Environments IDM-Mobil} are shown in Figure \ref{fig:IDM_env_r} and Figure \ref{fig:IDM_env_r_10}. In Figure \ref{fig:IDM_env_r}, we show the before and after adaptation of PEARL and MAML in the logarithmic axis, and in Figure \ref{fig:IDM_env_r_10}, we offer the last ten iterations of the MAML and PEAR training curve. We can have a similar conclusion that the PEARL method is much more data-efficient than the MAML method. Moreover, from Figure \ref{fig:IDM_env_r_10}, we can see that the PEARL agent shows good adaptation that the after adaptation reward is much higher than the before adaptation reward. Since the reward design of the IDM-Mobil is different from the attacker’s environment, there is no intuitive crash line. Therefore, we only summarize the crash rate in Table \ref{table:crash_rate_IDM} in Section \ref{Evaluation Results}.

As can be seen from the training curves, the trained MAML agent shows almost no adaptation while the PEARL agent can adapt efficiently. The reason is, in PEARL, the stochastic encoder takes full path data as input (actually $(s, a, s', r)$ tuples), so the trained PEARL agent would be able to capture the features of different environments with different transition probabilities. In comparison, the trained MAML agent in our experience is not sensitive to varying transition probabilities of different environments/tasks, which seems logical since the integrated inner/outer loops of MAML policy update Equation (\ref{Eq:MAML}) involves only the reward function.

\subsection{Evaluation Results}
\label{Evaluation Results}
In this section, we evaluate the trained agent with random tasks sampled from each distribution of environments. The evaluation results of the attackers environments are shown in Figure \ref{fig:attacker_env_r_eval} and the results of the IDM-Mobil environments are shown in Figure \ref{fig:IDM_env_r_eval}. We compared the PEARL approach and the MAML with the fine-tune approach in which we keep training the policy in a new environment. The x-axis is how much data we provide for the adaptation step. Each trajectory is at most 200 time steps if not crash. As you can see, after collecting two trajectories of data (at most 400 data points), the PEARL can adapt to new environments well in both distributions of environments. However, the MAML and fine-tune methods do not show improvement even with ten trajectories of data. This is because that the collected data in the new environment are not useful for the MAML agent and fine-tune agent to update its policies.

Next, we report the different agents’ crash rates during evaluation in Table \ref{table:crash_rate_att} and Table \ref{table:crash_rate_IDM} for the attacker environments and IDM-Mobil environments, respectively. All the methods are evaluated in random environments. On the leftmost column, we have the benchmark policy from Section \ref{Baselines}. The crash rate of the trained agent in the original environment is very low. However, when we test it in random environments, the crash rate increases significantly in both setups. For the fine-tune approach, the result shows that the agent cannot adapt to new environments with limited data, so the crash rate in new environments is around the same level for both setups. 

In the attacker environments, as shown in Table \ref{table:crash_rate_att}, the MAML agent keeps getting worse and worse, given the data. This due to insufficient exploration during the adaptation. Meanwhile, the PEARL agent can adapt to a new environment quickly with limited data. The crash rate of the PEARL agent reaches a very small number given 10 trajectories of data, which can compare to the benchmark's crash rate in the original environment. 


\begin{table}[ht]
	\caption{Crash Rate with Different Numbers of Data in the Attacker Environments}
	\centering
	\label{table:crash_rate_att}
	\begin{tabular}[h]{m{0.16\linewidth}|M{0.18\linewidth}|M{0.18\linewidth}|M{0.12\linewidth}|M{0.12\linewidth}}
		\hline
		Crash Rate & Benchmark & Fine Tune & MAML & PEARL \\
		\hline
		\hline
		Orig. task & $\sim0.001\%$ &$\sim0.001\%$&-&-\\
		\hline
		No adapt &\multirow{6}{*}{$17.8\%$} & $13.2\%$ & $19.4\%$ & $59.1\%$\\ 
		\cline{1-1} \cline{3-5}
		1 traj. & & $13.4\%$ & $22.5\%$ & $7.3\%$\\
		\cline{1-1} \cline{3-5}
		2 traj. & & $14.1\%$ & $24.9\%$ & $0.099\%$\\
		\cline{1-1} \cline{3-5}
		3 traj. & & $13.3\%$ & $27.3\%$ & $0.077\%$\\
		\cline{1-1} \cline{3-5}
		5 traj. & & $13.7\%$ & $31.4\%$ & $0.015\%$\\
		\cline{1-1} \cline{3-5}
		10 traj. & & $13.8\%$ & $36.9\%$ & $0.0062\%$\\
		\hline
	\end{tabular}
\end{table}

In IDM-Mobil environments, the MAML agent has better crash rates with more and more given data. However, the improvement still not significant enough compared to the PEARL agent. As can be seen from Table \ref{table:crash_rate_IDM}, the PEARL can adapt to a new environment quickly with limited data. The crash rate of the PEARL is comparable to the benchmark's crash rate in the original environment. Since in the IDM-Mobil environments, there is no short-horizon safety check, the benchmark crash rate is higher than the attacker environment. Moreover, in the IDM-Mobil environments, the agent controls the steering angle and the acceleration directly without any robust lower level controller. This causes a higher crash rate compared to the attacker environment. The crash rate results show that the PEARL trained agent can achieve the benchmark level crash rate with only ten trajectories of data in both setups. 

\begin{table}[ht]
	\caption{Crash Rate with Different Numbers of Data in the IDM-Mobil Environments}
	\label{table:crash_rate_IDM}
	\begin{tabular}[h]{m{0.16\linewidth}|M{0.18\linewidth}|M{0.18\linewidth}|M{0.12\linewidth}|M{0.12\linewidth}}
		\hline
		Crash Rate & Benchmark & Fine Tune & MAML & PEARL \\
		\hline
		\hline
		Orig. task & $\sim4\%$ &$\sim4\%$&-&-\\
		\hline
		No adapt &\multirow{6}{*}{$50.3\%$} & $52.6\%$ & $50.4\%$ & $60.3\%$\\ 
		\cline{1-1} \cline{3-5}
		1 traj. & & $51.9\%$ & $32.6\%$ & $26.7\%$\\
		\cline{1-1} \cline{3-5}
		2 traj. & & $49.5\%$ & $36.7\%$ & $18.1\%$\\
		\cline{1-1} \cline{3-5}
		3 traj. & & $49.8\%$ & $34.5\%$ & $12.7\%$\\
		\cline{1-1} \cline{3-5}
		5 traj. & & $48.2\%$ & $32.2\%$ & $10.5\%$\\
		\cline{1-1} \cline{3-5}
		10 traj. & & $47.6\%$ & $31.8\%$ & $5.2\%$\\
		\hline
	\end{tabular}
\end{table}
\section{CONCLUSIONS}
\label{CONCLUSIONS}

In this work we showed that, with all things being equal, solving the multi-MDPs problem offers significant adaptability for the resulted discretionary lane change policy under varying traffic cultures. This is important since surrounding drivers can behave differently at a different time of the day or in different weather conditions, and a trained policy under normal traffic behavior can be brittle when experiencing adversarial vehicles that can exploit the weakness of a fixed automated driving policy \cite{Zhang_2020_CVPR_Workshops}.

To observe how well an AV can adapt to an unseen traffic environment, two types of distribution variation in environments were designed, i.e., varying density of adversarial vehicles and varying mixture in surrounding vehicles' social behaviors. We witnessed that both our MRL (with MAML and PEARL approaches) enabled the AV to be a quick learner of the encountered new traffic behavior when compared to the baseline of a classic RL that fine-tuned its policy with the same amount of trajectory data.  Within ten data trajectories, MRL driving policy quickly adapted to the unseen traffic culture with the new MDP transition probability, and reduced crash rate significantly when compared to the baseline RL driving policy.



\section*{ACKNOWLEDGMENT}

We would like to thank the authors in paper \cite{nageshrao2019autonomous} for making the our bench mark comparison possible. 



\begin{CJK*}{UTF8}{gbsn}
\balance
\bibliography{main.bib}{}

\begin{thebibliography}{10}

\bibitem{justesen2019deep}
N.~Justesen, P.~Bontrager, J.~Togelius, and S.~Risi, ``Deep learning for video
  game playing,'' {\em IEEE Transactions on Games}, vol.~12, no.~1, pp.~1--20,
  2019.

\bibitem{kober2013reinforcement}
J.~Kober, J.~A. Bagnell, and J.~Peters, ``Reinforcement learning in robotics: A
  survey,'' {\em The International Journal of Robotics Research}, vol.~32,
  no.~11, pp.~1238--1274, 2013.

\bibitem{melis2017state}
G.~Melis, C.~Dyer, and P.~Blunsom, ``On the state of the art of evaluation in
  neural language models,'' {\em arXiv preprint arXiv:1707.05589}, 2017.

\bibitem{silver2013concurrent}
D.~Silver, L.~Newnham, D.~Barker, S.~Weller, and J.~McFall, ``Concurrent
  reinforcement learning from customer interactions,'' in {\em International
  conference on machine learning}, pp.~924--932, PMLR, 2013.

\bibitem{miotto2018deep}
R.~Miotto, F.~Wang, S.~Wang, X.~Jiang, and J.~T. Dudley, ``Deep learning for
  healthcare: review, opportunities and challenges,'' {\em Briefings in
  bioinformatics}, vol.~19, no.~6, pp.~1236--1246, 2018.

\bibitem{bazzan2013introduction}
A.~L. Bazzan and F.~Kl{\"u}gl, ``Introduction to intelligent systems in traffic
  and transportation,'' {\em Synthesis Lectures on Artificial Intelligence and
  Machine Learning}, vol.~7, no.~3, pp.~1--137, 2013.

\bibitem{mnih2015human}
V.~Mnih, K.~Kavukcuoglu, D.~Silver, A.~A. Rusu, J.~Veness, M.~G. Bellemare,
  A.~Graves, M.~Riedmiller, A.~K. Fidjeland, G.~Ostrovski, {\em et~al.},
  ``Human-level control through deep reinforcement learning,'' {\em nature},
  vol.~518, no.~7540, pp.~529--533, 2015.

\bibitem{silver2016mastering}
D.~Silver, A.~Huang, C.~J. Maddison, A.~Guez, L.~Sifre, G.~Van Den~Driessche,
  J.~Schrittwieser, I.~Antonoglou, V.~Panneershelvam, M.~Lanctot, {\em et~al.},
  ``Mastering the game of go with deep neural networks and tree search,'' {\em
  nature}, vol.~529, no.~7587, pp.~484--489, 2016.

\bibitem{li2017deep}
Y.~Li, ``Deep reinforcement learning: An overview,'' {\em arXiv preprint
  arXiv:1701.07274}, 2017.

\bibitem{ng2000algorithms}
A.~Y. Ng, S.~J. Russell, {\em et~al.}, ``Algorithms for inverse reinforcement
  learning.,'' in {\em Icml}, vol.~1, p.~2, 2000.

\bibitem{eysenbach2018diversity}
B.~Eysenbach, A.~Gupta, J.~Ibarz, and S.~Levine, ``Diversity is all you need:
  Learning skills without a reward function,'' {\em arXiv preprint
  arXiv:1802.06070}, 2018.

\bibitem{wen2020safe}
L.~Wen, J.~Duan, S.~E. Li, S.~Xu, and H.~Peng, ``Safe reinforcement learning
  for autonomous vehicles through parallel constrained policy optimization,''
  in {\em 2020 IEEE 23rd International Conference on Intelligent Transportation
  Systems (ITSC)}, pp.~1--7, IEEE, 2020.

\bibitem{nageshrao2019autonomous}
S.~Nageshrao, H.~E. Tseng, and D.~Filev, ``Autonomous highway driving using
  deep reinforcement learning,'' in {\em 2019 IEEE International Conference on
  Systems, Man and Cybernetics (SMC)}, pp.~2326--2331, IEEE, 2019.

\bibitem{nageshrao2020vehicle}
S.~Nageshrao, H.~E. Tseng, D.~P. Filev, R.~L. Baker, C.~Cruise, L.~Daehler,
  S.~Mohan, and A.~Kusari, ``Vehicle adaptive learning.'' U.S. 10733510, August
  2020.

\bibitem{chow2018lyapunov}
Y.~Chow, O.~Nachum, E.~Duenez-Guzman, and M.~Ghavamzadeh, ``A lyapunov-based
  approach to safe reinforcement learning,'' {\em arXiv preprint
  arXiv:1805.07708}, 2018.

\bibitem{wachi2018safe}
A.~Wachi, Y.~Sui, Y.~Yue, and M.~Ono, ``Safe exploration and optimization of
  constrained mdps using gaussian processes,'' in {\em Proceedings of the AAAI
  Conference on Artificial Intelligence}, vol.~32, 2018.

\bibitem{dulac2019challenges}
G.~Dulac-Arnold, D.~Mankowitz, and T.~Hester, ``Challenges of real-world
  reinforcement learning,'' {\em arXiv preprint arXiv:1904.12901}, 2019.

\bibitem{jaafra2019robust}
Y.~Jaafra, J.~L. Laurent, A.~Deruyver, and M.~S. Naceur, ``Robust reinforcement
  learning for autonomous driving,'' {\em APIA}, p.~52, 2019.

\bibitem{nagabandi2018learning}
A.~Nagabandi, I.~Clavera, S.~Liu, R.~S. Fearing, P.~Abbeel, S.~Levine, and
  C.~Finn, ``Learning to adapt in dynamic, real-world environments through
  meta-reinforcement learning,'' {\em arXiv preprint arXiv:1803.11347}, 2018.

\bibitem{finn2017one}
C.~Finn, T.~Yu, T.~Zhang, P.~Abbeel, and S.~Levine, ``One-shot visual imitation
  learning via meta-learning,'' in {\em Conference on Robot Learning},
  pp.~357--368, PMLR, 2017.

\bibitem{yu2018one}
T.~Yu, C.~Finn, A.~Xie, S.~Dasari, T.~Zhang, P.~Abbeel, and S.~Levine,
  ``One-shot imitation from observing humans via domain-adaptive
  meta-learning,'' {\em arXiv preprint arXiv:1802.01557}, 2018.

\bibitem{duan2016rl}
Y.~Duan, J.~Schulman, X.~Chen, P.~L. Bartlett, I.~Sutskever, and P.~Abbeel,
  ``$\text{RL}^2$: Fast reinforcement learning via slow reinforcement
  learning,'' {\em arXiv preprint arXiv:1611.02779}, 2016.

\bibitem{wang2016learning}
J.~X. Wang, Z.~Kurth-Nelson, D.~Tirumala, H.~Soyer, J.~Z. Leibo, R.~Munos,
  C.~Blundell, D.~Kumaran, and M.~Botvinick, ``Learning to reinforcement
  learn,'' {\em arXiv preprint arXiv:1611.05763}, 2016.

\bibitem{heess2015memory}
N.~Heess, J.~J. Hunt, T.~P. Lillicrap, and D.~Silver, ``Memory-based control
  with recurrent neural networks,'' {\em arXiv preprint arXiv:1512.04455},
  2015.

\bibitem{finn2017model}
C.~Finn, P.~Abbeel, and S.~Levine, ``Model-agnostic meta-learning for fast
  adaptation of deep networks,'' in {\em International Conference on Machine
  Learning}, pp.~1126--1135, PMLR, 2017.

\bibitem{rothfuss2018promp}
J.~Rothfuss, D.~Lee, I.~Clavera, T.~Asfour, and P.~Abbeel, ``Promp: Proximal
  meta-policy search,'' {\em arXiv preprint arXiv:1810.06784}, 2018.

\bibitem{gupta2018meta}
A.~Gupta, R.~Mendonca, Y.~Liu, P.~Abbeel, and S.~Levine, ``Meta-reinforcement
  learning of structured exploration strategies,'' {\em arXiv preprint
  arXiv:1802.07245}, 2018.

\bibitem{rakelly2019efficient}
K.~Rakelly, A.~Zhou, C.~Finn, S.~Levine, and D.~Quillen, ``Efficient off-policy
  meta-reinforcement learning via probabilistic context variables,'' in {\em
  International conference on machine learning}, pp.~5331--5340, PMLR, 2019.

\bibitem{haarnoja2018soft}
T.~Haarnoja, A.~Zhou, P.~Abbeel, and S.~Levine, ``Soft actor-critic: Off-policy
  maximum entropy deep reinforcement learning with a stochastic actor,'' in
  {\em International Conference on Machine Learning}, pp.~1861--1870, PMLR,
  2018.

\bibitem{AVstatistics}
J.~Stewart, ``Why people keep rear-ending self-driving cars.''
  \url{https://www.wired.com/story/self-driving-car-crashes-rear-endings-why-charts-statistics/},
  2018.
\newblock Accessed: 04.03.2021.

\bibitem{Zhang_2020_CVPR_Workshops}
S.~Zhang, H.~Peng, S.~Nageshrao, and H.~E. Tseng, ``Generating socially
  acceptable perturbations for efficient evaluation of autonomous vehicles,''
  in {\em Proceedings of the IEEE/CVF Conference on Computer Vision and Pattern
  Recognition (CVPR) Workshops}, June 2020.

\bibitem{highway-env}
E.~Leurent, ``An environment for autonomous driving decision-making.''
  \url{https://github.com/eleurent/highway-env}, 2018.

\bibitem{schwarting2019social}
W.~Schwarting, A.~Pierson, J.~Alonso-Mora, S.~Karaman, and D.~Rus, ``Social
  behavior for autonomous vehicles,'' {\em Proceedings of the National Academy
  of Sciences}, vol.~116, no.~50, pp.~24972--24978, 2019.

\bibitem{akiba2019optuna}
T.~Akiba, S.~Sano, T.~Yanase, T.~Ohta, and M.~Koyama, ``Optuna: A
  next-generation hyperparameter optimization framework,'' in {\em Proceedings
  of the 25th ACM SIGKDD international conference on knowledge discovery \&
  data mining}, pp.~2623--2631, 2019.

\bibitem{schulman2015trust}
J.~Schulman, S.~Levine, P.~Abbeel, M.~Jordan, and P.~Moritz, ``Trust region
  policy optimization,'' in {\em International conference on machine learning},
  pp.~1889--1897, PMLR, 2015.

\bibitem{zhang2019discretionary}
S.~Zhang, H.~Peng, S.~Nageshrao, and E.~Tseng, ``Discretionary lane change
  decision making using reinforcement learning with model-based exploration,''
  in {\em 2019 18th IEEE International Conference On Machine Learning And
  Applications (ICMLA)}, pp.~844--850, IEEE, 2019.

\end{thebibliography}
\bibliographystyle{ieeetr}
\end{CJK*}
\end{document}